\documentclass[]{spie}  

\usepackage{amsmath,amsfonts,amssymb}
\usepackage{graphicx}
\usepackage[colorlinks=true, allcolors=blue]{hyperref}

\title{Predicting Alzheimer's Disease Progression Using rs-fMRI and a 
History-Aware Graph Neural Network}

\author[a]{Mahdi Moghaddami}
\author[a]{Mohammad-Reza Siadat}
\author[a]{Austin T. Toma}
\author[a]{Connor P. Laming}
\author[a]{Huirong Fu}
\affil[a]{Oakland University, 502 Meadow Brook Road, Rochester, 
MI 48309, USA}

\authorinfo{Send correspondence to M.M.: moghaddami@oakland.edu}

\begin{document}
\maketitle

\begin{abstract}
    Alzheimer's disease (AD) is a neurodegenerative disorder that affects
    more than seven million people in the United States alone. AD currently
    has no cure, but there are ways to potentially slow its progression if
    caught early enough. In this study, we propose a graph neural network
    (GNN)-based model for predicting whether a subject will transition to a
    more severe stage of cognitive impairment at their next clinical visit.
    We consider three stages of cognitive impairment in order of severity:
    cognitively normal (CN), mild cognitive impairment (MCI), and AD. We use
    functional connectivity graphs derived from resting-state functional
    magnetic resonance imaging (rs-fMRI) scans of 303 subjects, each with
    a different number of visits. Our GNN-based model incorporates a recurrent
    neural network (RNN) block, enabling it to process data from the
    subject's entire visit history. It can also work with irregular time
    gaps between visits by incorporating visit distance information into
    our input features. Our model demonstrates robust predictive performance,
    even with missing visits in the subjects' visit histories. It achieves
    an accuracy of 82.9\%, with an especially impressive accuracy of 68.8\% on
    CN to MCI conversions - a task that poses a substantial challenge in the
    field. Our results highlight the effectiveness of rs-fMRI in predicting
    the onset of MCI or AD and, in conjunction with other modalities, could
    offer a viable method for enabling timely interventions to slow the
    progression of cognitive impairment.
\end{abstract}

\section{Introduction}
Alzheimer's disease (AD) is a progressive neurodegenerative
disorder and the most common cause of dementia worldwide\cite{SCHELTENS2016505},
resulting in substantial clinical, social, and economic burden to patients and society as a
whole.
AD is characterized by gradual cognitive decline that unfolds over several years,
progressing from a cognitively normal (CN) state through mild cognitive impairment
(MCI) to advanced dementia. Although no cure currently exists, early detection
allows for timely interventions that may slow disease progression and improve
quality of life\cite{lancet}. Consequently, accurately predicting stage
transitions remains a critical research objective.

Given the importance of early detection, neuroimaging has played a central role
in advancing the understanding of AD-related brain
changes\cite{Jack2008TheAD,Mulumba2025}. In particular, resting-state functional
magnetic resonance imaging (rs-fMRI) has emerged as a robust noninvasive modality
for investigating functional connectivity (FC) patterns in the
brain\cite{GRANZIERA2015917}. FC quantifies the temporal correlation
of blood-oxygen-level-dependent (BOLD) signals between spatially distinct brain
regions, revealing intrinsic functional correlation.
Alterations in functional brain networks have been shown to correlate with
cognitive decline and disease progression\cite{Mulumba2025}, establishing rs-fMRI
a promising data source for predictive modeling. Nevertheless, effectively utilizing
rs-fMRI data remains difficult due to its high dimensionality, complex network
form, and subject- and time-point-specific variability.

Recent advances in machine learning, particularly deep learning, have enabled
more sophisticated analyses of neuroimaging data. Graph
neural networks (GNNs) are especially well-suited for modeling
functional connectivity since FC matrices naturally form
graph-structured data where brain regions are represented as nodes (vertices)
and functional relationships as edges\cite{Hamilton2017InductiveRL}.
Previous studies, such as BrainGNN\cite{Li2020BrainGNNIB} have shown the
effectiveness of GNN-based approaches for brain network analysis and AD diagnosis,
indicating that graph convolutional layers can effectively extract complex
spatial relationships in functional connectivity networks. However,
most existing GNN methods focus on cross-sectional data and do not fully
utilize the longitudinal aspects of disease progression\cite{moghaddami25a}.

Longitudinal prediction poses additional challenges, including
irregular time intervals between clinical visits, missing imaging
data, and varying lengths of visit history across subjects\cite{moghaddami25a}.
These factors complicate the modeling of temporal processes and restrict the
applicability of standard sequential or graph-based models.
Recurrent neural networks (RNNs) such as long short-term memory
(LSTM)\cite{hochreiter1997long} and gated recurrent units (GRU)\cite{Cho2014LearningPR}
have shown potential in modeling temporal sequences, yet their application to
longitudinal neuroimaging data with irregular sampling remains difficult.

In this study, we tackle these challenges by introducing a
history-aware graph neural network (HA-GNN) model, which integrates
rs-fMRI-derived functional connectivity graphs with longitudinal
visit information. Our strategy combines a GNN for spatial brain network
representation with an RNN for temporal representation,
enabling the capture of both organizational characteristics of functional brain
networks and their temporal variation. To accommodates irregular visit spacing and
incomplete visit histories, we explicitly embed temporal intervals between
consecutive visits.
We evaluate the proposed method using rs-fMRI data from the Alzheimer's Disease
Neuroimaging Initiative (ADNI)\cite{Jack2008TheAD} dataset and demonstrate that
using longitudinal rs-fMRI data within a history-aware graph learning
framework can boost the prediction of cognitive impairment
progression, particularly in identifying
subjects who transition to further severe disease stages.

\section{Method}
\label{sec:method}

\subsection{Data}
We use rs-fMRI and structural MRI (sMRI) scans provided by the ADNI dataset.
This data is available publicly at \url{https://ida.loni.usc.edu/}.
Rs-fMRI is a type of brain scan that shows which areas of the brain are active in
the absence of tasks or external stimuli\cite{GRANZIERA2015917}.


We focus on subjects from the Alzheimer's disease prediction of longitudinal
evolution (TADPOLE) dataset\cite{Marinescu2018TADPOLECP} who have at least two
rs-fMRI scan and at least one sMRI scan in the ADNI dataset.
At least two visits are required since we want to make a prediction
about a future date and having only one visit would not provide a
ground truth value for our prediction. We remove ``reverters'', i.e.,
subjects that convert to a less severe stage in their last visit (AD to MCI,
MCI to CN, or AD to CN).
This leaves us with 303 participants from the ADNI dataset meeting these
criteria, each one having a different number of available rs-fMRI
scans taken at different times.

\begin{definition}[Converter Subject]
    \label{def:converter_subject}
    A subject that transitions to a more severe stage of cognitive impairment
    (CN to MCI, CN to AD, or MCI to AD) in their last visit.
\end{definition}

\begin{definition}[Stable Subject]
    \label{def:stable_subject}
    A subject with the same diagnosis in their last visit as their penultimate visit.
\end{definition}

\begin{figure} [h]
    \begin{center}
        \begin{tabular}{c} 
            \includegraphics[width=0.6\linewidth]{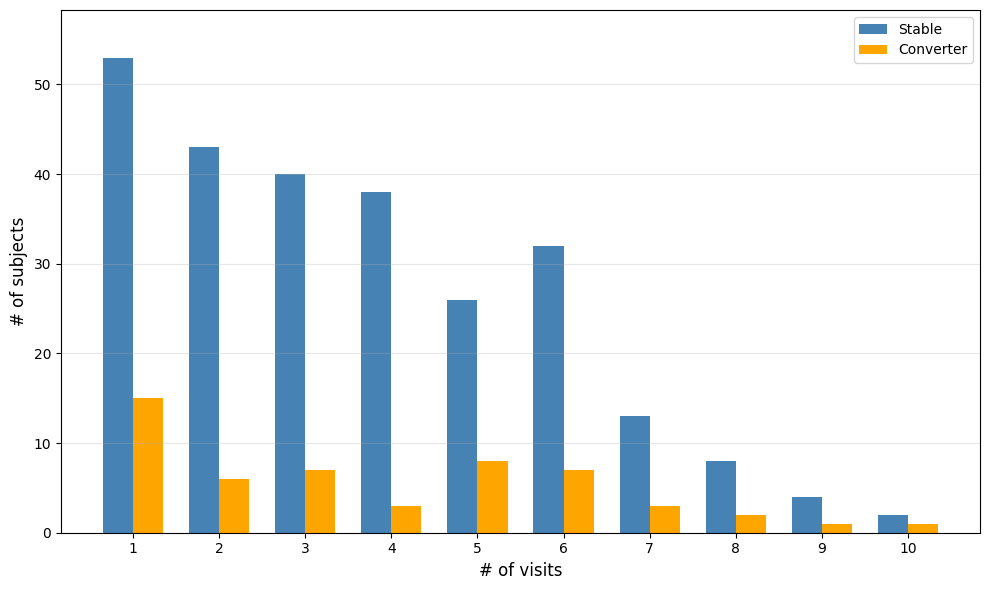}
        \end{tabular}
    \end{center}
    \caption[example]
    { \label{fig:data-groups}
        Distribution of subjects based on their number of available rs-fMRI scans.
        The blue bars represent stable subjects, while the orange bars represent
        converter subjects. \vspace{0.5 cm}}
\end{figure}

Fig.~\ref{fig:data-groups} shows the distribution of 1089 rs-fMRI
scans in total for 303 subjects, with 894 scans from stable subjects and
195 scans from converter subjects. The distances between consecutive scans
for subjects are irregular, with a mean of 14.78 months.

\begin{figure} [h]
    \begin{center}
        \begin{tabular}{c} 
            \includegraphics[width=0.6\linewidth]{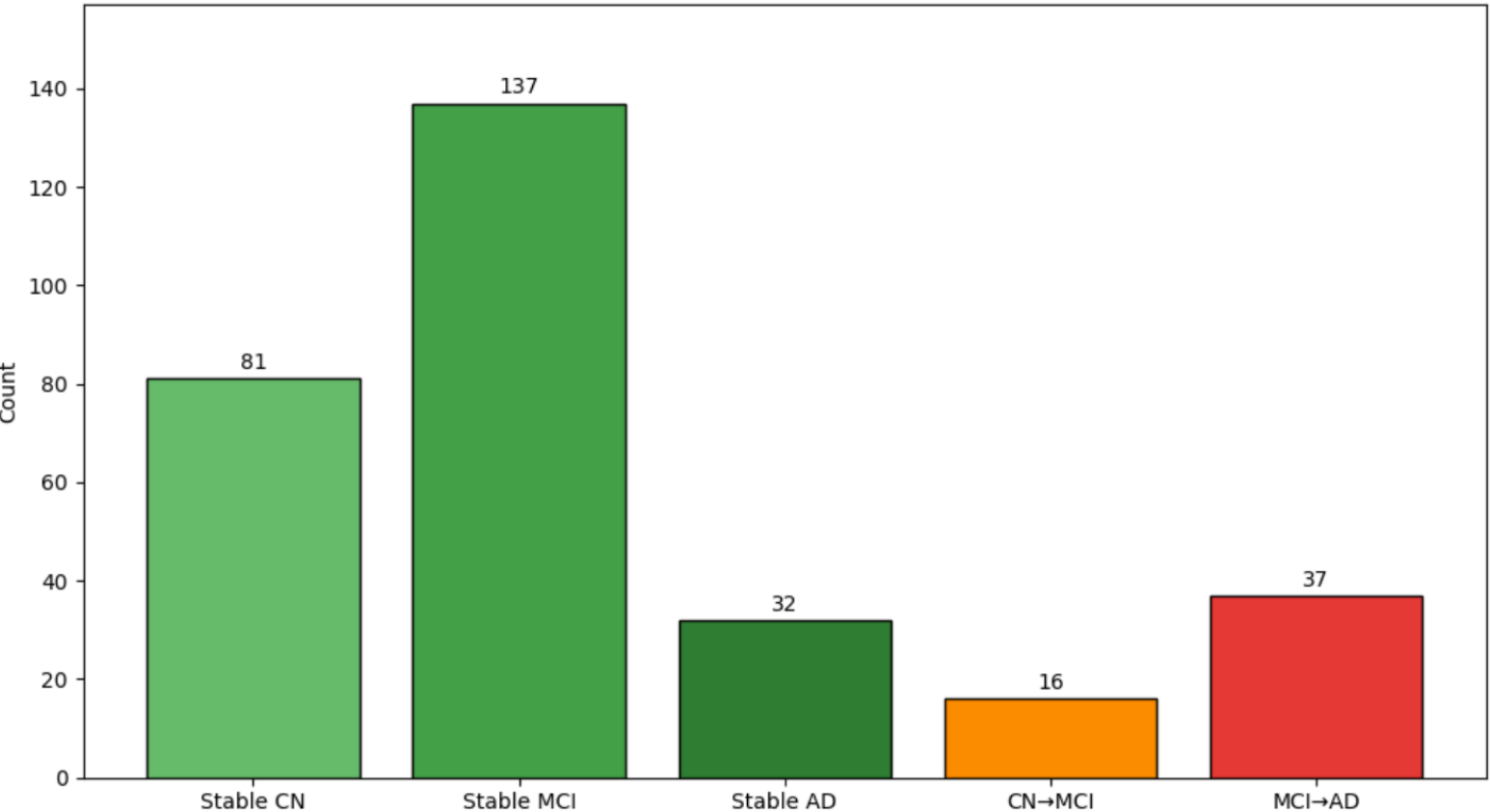}
        \end{tabular}
    \end{center}
    \caption[example]
    { \label{fig:data-classes}
        Distribution of subjects based on their diagnosis in their last visit}
\end{figure}

Expectedly, the number of scans from stable subjects is much higher than
that of converter subjects. Also, subjects with histories longer than seven visits
are rare. We have 250 stable subjects and 53 converter subjects, as shown in
Fig.~\ref{fig:data-classes}. There are no subjects converting from CN to AD in
their last visit.

\subsection{Preprocessing}
FMRI scans need to be preprocessed before we can use them for our task to
account for external sources of nuisance imaging artifacts \cite{Esteban2018}.
The FMRIPrep is a pre-processing pipeline that has a multitude of features designed
to ready the fMRI data in order to be used for analysis\cite{Esteban2018}.

We apply the fMRIPrep preprocessing pipeline with default parameters, which
includes the following steps: (1) anatomical preprocessing with skull stripping, (2) brain tissue segmentation into cerebrospinal fluid (CSF),
white matter (WM), and gray matter (GM), (3) spatial normalization to MNI152NLin2009cAsym
standard space, (4) functional preprocessing including head motion correction using
volume-to-volume rigid body registration, (5) slice-timing correction to account for
interleaved acquisition, (6) co-registration of functional images to the corresponding anatomical scan,
and (7) resampling to the target output space with motion parameters extracted for
subsequent confound regression.

As fMRIPrep requires its input images to be in NIfTI
format and stored in the Brain Imaging Data Structure
(BIDS)\cite{Gorgolewski2016TheBI} directory structure, we convert the
downloaded DICOM files to NIfTI and then organize them in the BIDS format. We
then run fMRIPrep on the BIDS directory containing structural and functional scans
for all subjects to get the preprocessed rs-fMRI scans.

\subsection{Functional Connectivity Matrix Computation}

The BOLD signals are valuable information derived from fMRI volumes.
Due to oxygen in the brain being slightly magnetic, we are able to get the
amount of oxygen in each part of the brain at different times
\cite{Mulumba2025}. This can be used to find correlations between the activities of
regions of the brain and detect irregularities in these correlations and overall
function.

One way to extract information from fMRI scans is to obtain the functional
connectivity matrix. To do this, we need to first parcellate the brain into
regions of interest (ROIs) to be used as nodes to relate to each other. We utilize
the Nilearn library and the Schaefer atlas\cite{Schaefer2017} to
parcellate the brain into 100 ROIs. We then extract the time series from each
ROI by averaging the signals from all pixels in each ROI and compute the functional
connectivity matrix using Pearson correlation. The FC matrix is a square matrix
where each element represents the correlation between the time series of two ROIs.

\subsection{Model}
\begin{figure} [ht]
    \begin{center}
        \begin{tabular}{c} 
            \includegraphics[width=0.8\linewidth]{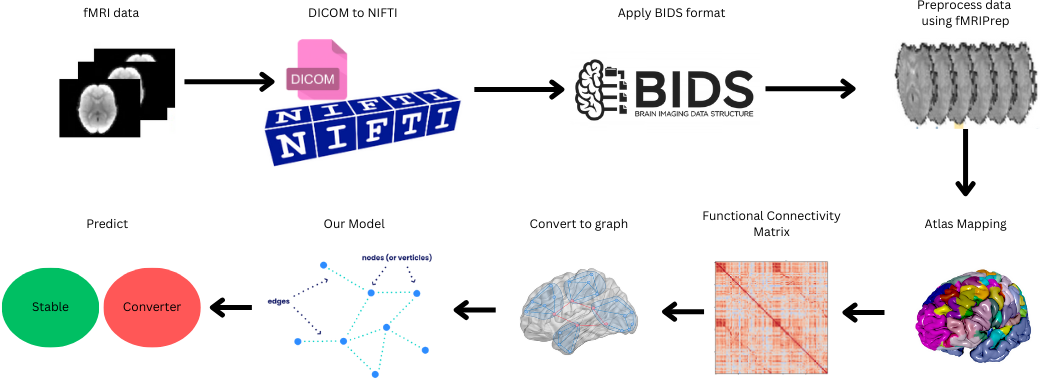}
        \end{tabular}
    \end{center}
    \caption[example]
    { \label{fig:entire-process}
        Overview of our workflow, from raw rs-fMRI scans to the binary prediction}
\end{figure}

We consider a binary classification task where we want to predict whether a
subject will convert to a more severe stage of cognitive impairment in their
last visit. Similarly to
Ref.~\citenum{moghaddami25a}, the input to
our model is the first $n$ visits from the visit history
$v=(v_1, v_2, \ldots, v_{n+1})$ of a subject, where $v_i$ is the FC matrix
of the $i$-th visit and $n+1$ is the total number of visits. The output is `stable'
if $d_n=d_{n+1}$, where $d_i\in$\{CN, MCI, AD\} is the diagnosis of the subject in
the $i$-th visit, and `converter' otherwise. Fig.~\ref{fig:entire-process} shows
all the steps taken to get to the final prediction from rs-fMRI scans.

\begin{figure} [ht]
    \begin{center}
        \begin{tabular}{c} 
            \includegraphics[width=0.975\linewidth]{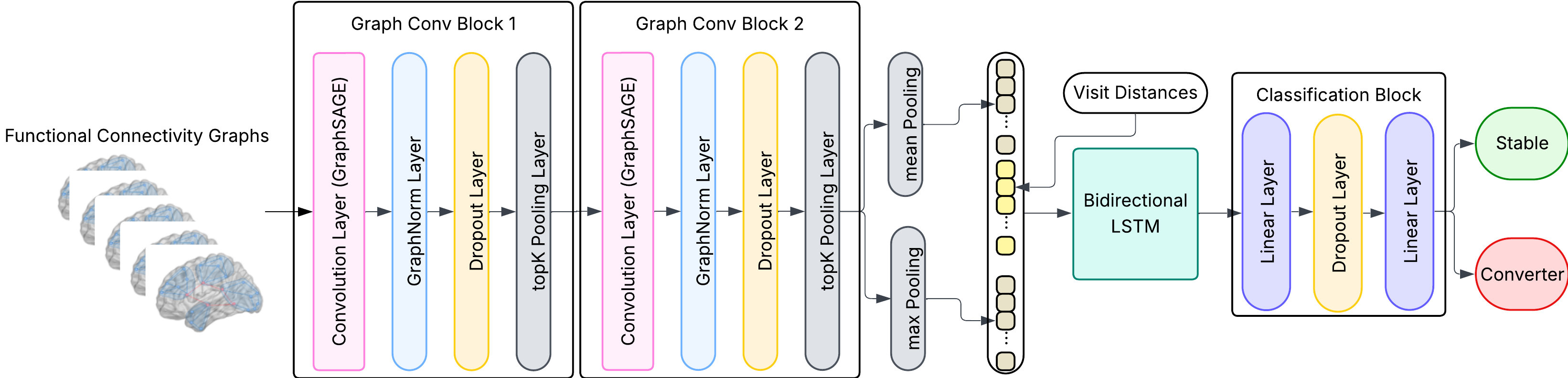}
        \end{tabular}
    \end{center}
    \caption[model]
    { \label{fig:model}
        Architecture of our HA-GNN model}
\end{figure}

In this work, we propose a history-aware graph neural network (GNN) model
combining two main components: a graph convolutional network (GCN) and an RNN.
The GCN is loosely based on the
BrainGNN\cite{Li2020BrainGNNIB} model, having two convolutional blocks. Unlike
BrainGNN, our model is capable of working with multiple data points. Each block
consists of a GraphSAGE layer\cite{Hamilton2017InductiveRL}, a graph normalization
layer\cite{Cai2020GraphNormAP}, a dropout layer, and A topK pooling
layer\cite{Li2020BrainGNNIB}. The RNN component can be any choice of RNN, such as
a long short-term memory (LSTM)\cite{hochreiter1997long} or a gated
recurrent unit (GRU)\cite{Cho2014LearningPR}. Fig.~\ref{fig:model} shows the
architecture of our proposed model in detail.

Between the GCN and RNN components, we concatenate the distances between
consecutive visits for the input subject to the rest of our feature vector. This
is done to provide the model with information about the time intervals between
visits and how far ahead it needs to predict.

\subsection{Training and Evaluation}
\begin{figure} [ht]
    \begin{center}
        \begin{tabular}{c} 
            \includegraphics[width=0.6\linewidth]{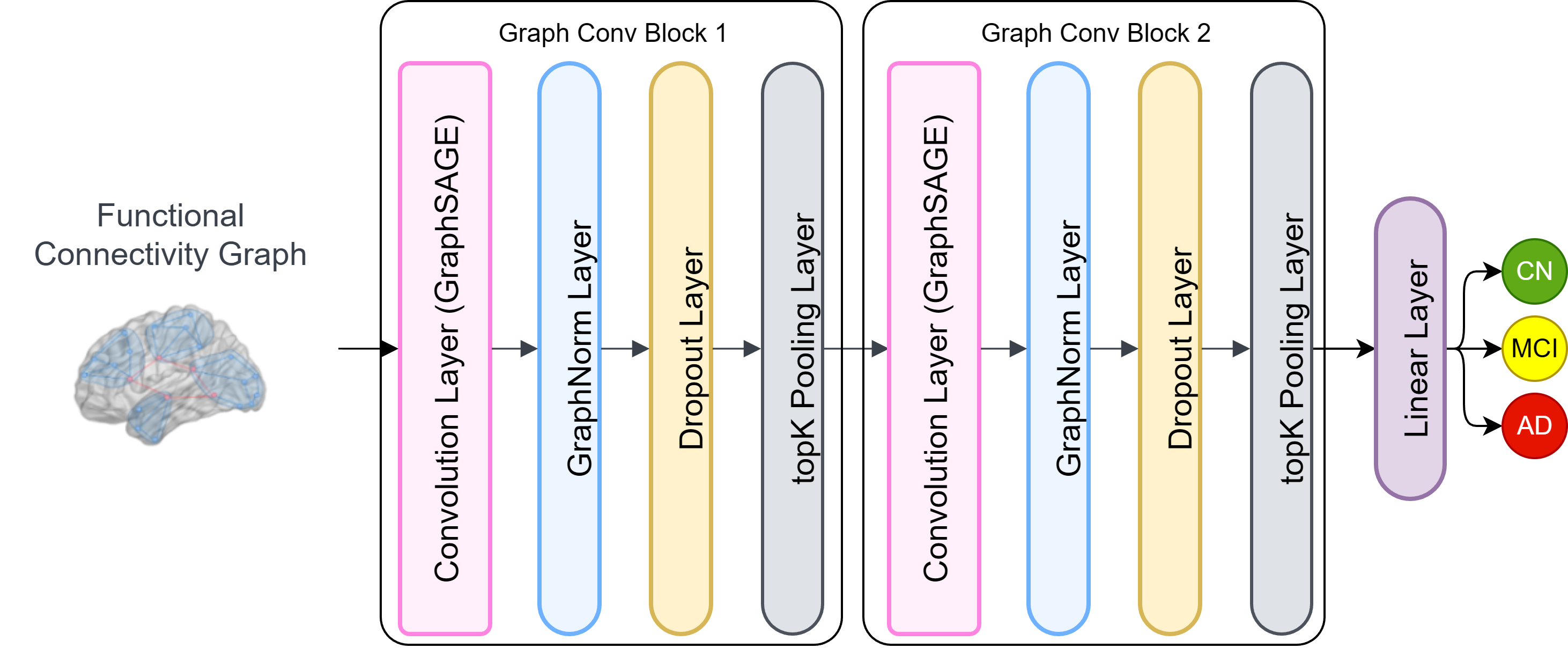}
        \end{tabular}
    \end{center}
    \caption[pretraining model]
    { \label{fig:pretraining}
        Architecture of the GCN pretraining model}
\end{figure}

We use a random 20\% split of our data to pretrain our GCN component. The
pretraining is a 3-class classification task where the model predicts the
diagnosis of a subject (CN, MCI, or AD) in a given visit. Unlike our main task, during
pretraining, the input consists of a single visit instead of the entire visit
history of a subject. Fig.~\ref{fig:pretraining} shows the architecture of the
pretraining model.

Once pretraining is done, we use 5-fold cross-validation to train and evaluate
the model on the main prediction task. We use Bayesian
optimization\cite{BayesianOptimization} for hyperparameter tuning and utilize
focal loss\cite{Lin2017FocalLF}($\alpha=0.9, \gamma=3$) to address the
class imbalance present in our data.

To evaluate our model, we use balanced accuracy (BA), which is
defined as the average of sensitivity and specificity. We also report the area
under the receiver operating characteristic curve (AUC-ROC)\cite{Bradley1997TheUO}
and raw accuracy scores.

\section{Results}
\label{sec:results}

All results are reported as mean $\pm$ standard deviation across the 5-folds.
The results of the classification task are shown in Table~\ref{tab:results}.
CN to MCI and MCI to AD columns represent the accuracy of the model in identifying
CN to MCI and MCI to AD conversions, respectively. The last row shows the results
of the HA-GNN (GRU) model without the GCN pretraining step.

\begin{table}[ht]
    \centering
    \caption{Classification results of the proposed model}
    \label{tab:results}
    \begin{tabular}{|c|c|c|c|c|c|}
        \hline
        Model           & Acc.                       & AUC-ROC                    & BA                         & CN to MCI      & MCI to AD      \\
        \hline
        HA-GNN (LSTM)   & \textbf{0.829 $\pm$ 0.058} & \textbf{0.852 $\pm$ 0.065} & \textbf{0.771 $\pm$ 0.114} & \textbf{0.688} & \textbf{0.676} \\
        HA-GNN (RNN)    & 0.743 $\pm$ 0.038          & 0.771 $\pm$ 0.067          & 0.651 $\pm$ 0.114          & 0.5            & 0.514          \\
        HA-GNN (GRU)    & 0.733 $\pm$ 0.049          & 0.786 $\pm$ 0.062          & 0.704 $\pm$ 0.071          & 0.625          & 0.676          \\
        w/o pretraining & 0.723 $\pm$ 0.079          & 0.642 $\pm$ 0.095          & 0.594 $\pm$ 0.061          & 0.375          & 0.405          \\
        \hline
    \end{tabular}
\end{table}

Due to our novel methodology, comparisons with the existing literature are difficult.
However there are a few papers that are close to our task and methods.
Kim et al.\cite{Kim2021} provide an in-depth dive into certain ROIs and their
effectiveness in detecting the different stages of cognitive decline.
Grammenos et al.\cite{Grammenos2024} look into stable MCI (sMCI) subjects versus
converter MCI patients and detect the difference between stable and progressive
MCI (pMCI) patients. Their best-performing method of XGBoost resulted in an 86\% total
accuracy across both sMCI and pMCI subjects. Wang et al.
\cite{https://doi.org/10.48550/arxiv.2412.06212} also uses rs-fMRI from the ADNI
dataset and identifies subjects with AD. Their best-performing model for ADNI fMRI
data achieved an accuracy of 67.65\%.
\section{Discussion}

Our proposed history-aware graph neural network model achieved a balanced accuracy
of 77.1\% $\pm$ 11.4\% and an overall accuracy of 82.9\% $\pm$ 5.8\% in predicting
cognitive impairment progression using longitudinal rs-fMRI data. These results
demonstrate the feasibility of combining graph neural networks
with recurrent architectures to capture both spatial and temporal patterns in
functional brain connectivity. The model's performance is particularly notable given
the challenging characteristics of our dataset, including irregular visit intervals,
class imbalance, and the inherent difficulty of longitudinal prediction.

Moghaddami et al.\cite{moghaddami25a} employed a Transformer-based model on
structured tabular features, achieving a multi-class AUC of 0.890 $\pm$ 0.033
when predicting the diagnosis of subject in their next visit. While their
AUC is slightly higher than our 0.852 $\pm$ 0.065, their approach relied on
manually curated features, including cognitive test scores and imaging-derived
biomarkers. They also had a larger dataset. Our model operates directly on
functional connectivity matrices, preserving rich spatial information
in brain network topology. Our 68.8\% accuracy on CN to MCI conversion and 67.6\%
on MCI to AD conversion demonstrates effective detection of early-stage cognitive
changes.

Grammenos et al.\cite{Grammenos2024} focused specifically on distinguishing sMCI
subjects from pMCI patients using XGBoost on engineered features, achieving
86\% accuracy. While their accuracy appears higher, it is important to note that
their task was more narrowly defined and did not
address the full spectrum of cognitive transitions (CN to MCI, MCI to AD) that our
model handles. Additionally, they used structured features rather than raw
imaging data. Our converter detection rate of 69.4\% across all transition types
represents strong performance on a more complex and clinically diverse task.


Among the three RNN architectures tested, LSTM achieved the best performance across
all metrics, with a balanced accuracy of 77.1\%, substantially outperforming the
RNN (65.1\%) and slightly exceeding GRU (70.4\%). This result aligns with
the known advantages of LSTM in capturing long-term dependencies and mitigating
vanishing gradient problems\cite{hochreiter1997long}. The superior performance of
LSTM is particularly relevant in our context, where visit intervals span multiple
months or years, and the model must maintain information about earlier visits while
processing the temporal sequence.

The high standard deviation in balanced accuracy (11.4\%) across folds suggests
sensitivity to the particular distribution of converter subjects in each fold. This
is not surprising given the substantial class imbalance in our dataset (250 stable
subjects vs. 53 converter subjects) and highlights the importance of our use of
focal loss\cite{Lin2017FocalLF} to address this challenge. The lower standard
deviation in overall accuracy (5.8\%) reflects the model's consistent performance
on the majority stable class, while the variation in balanced accuracy reveals the
difficulty in consistently identifying the minority converter class across different
data splits.

Early detection at the CN to MCI transition
offers the greatest potential for intervention before substantial irreversible
neurodegeneration has occurred\cite{SCHELTENS2016505}. The fact that our model
achieves competitive performance at this critical stage suggests that rs-fMRI-derived
functional connectivity patterns contain predictive information even in the earliest
phases of cognitive decline.


Our two-stage training strategy, beginning with GCN pretraining on single-visit
diagnosis classification before training the full model on longitudinal prediction,
provides the network with a meaningful initialization. This pretraining helps the GCN
component learn representations of functional connectivity patterns
associated with different disease states, which the RNN component can then use
to learn temporal transitions.

\section{Limitations and Future Work}

Despite these promising results, several limitations warrant discussion. First, our
dataset size, while including 303 subjects and 1089 scans, remains relatively modest
for deep learning standards, and the imbalance between stable and converter subjects
limits our ability to fully train on rare transition patterns. The high variance in
balanced accuracy across folds suggests that larger datasets could improve
robustness. Future work could benefit from multi-site data aggregation or
data augmentation strategies tailored to rs-fMRI data.

Second, our model currently operates on functional connectivity alone and does not
leverage other available imaging modalities (structural MRI, diffusion MRI) or
non-imaging data (cognitive assessments, genetic information). Multimodal integration
has shown promise in related work\cite{https://doi.org/10.48550/arxiv.2412.06212}
and could potentially enhance our model's predictive power. However, such integration
must be carefully designed to preserve the advantages of our
current approach.

Third, the irregular and varying-length visit histories in our dataset, while
realistically reflecting clinical practice, pose challenges for model training and
evaluation. Our current approach handles this through embedding temporal intervals
between consecutive visits for each subject,
but more sophisticated methods for dealing with irregular temporal sampling could
be explored. Additionally, investigating the relationship between visit history
length and prediction accuracy could provide insights into the minimum number of data
points required for reliable prognostic predictions.

Fourth, while our model achieves respectable performance metrics, the lack of built-in
interpretability mechanisms limits its clinical utility. Future iterations should
incorporate attention mechanisms or other interpretability tools to identify which
brain regions and connections are most influential in driving predictions, as
demonstrated in Kim et al.\cite{Kim2021}. Such interpretability would not only
increase clinical trust in the model but could also provide neuroscientific insights
into the functional network changes that precede cognitive decline.

Finally, our model was trained and evaluated exclusively on ADNI data, which consists
primarily of North American participants recruited through specialized research
centers. Generalization to more diverse populations, different imaging protocols,
and real-world clinical settings remains to be validated. External validation on
independent cohorts is a necessity before clinical translation can
be considered.





\section{Conclusions}
\label{sec:conclusions}

In this study, we establish the effectiveness of using
resting-state functional MRI and our proposed history-aware
graph neural network model for predicting the progression of
cognitive impairment. Using focal loss and pretraining, our model
demonstrates the ability to predict converter subjects despite
data imbalance, missing visits, and irregular visit distances.
The results indicate that our model can effectively capture the
temporal dynamics of cognitive impairment progression. Our approach
offers a promising method for early detection and monitoring of
Alzheimer's disease which is crucial for timely intervention and
treatment. Subsequent research will focus on further improving the model's
performance and generalizability by employing additional data modalities
and examining the effects of history length on prediction accuracy.

\section*{ACKNOWLEDGMENTS}
This work was supported by the NSF grants \#CNS-2349663. This work
used Indiana JetStream2 GPU at Indiana University through allocation
NAIRR250048 from the Advanced Cyber infrastructure Coordination
Ecosystem:
Services \& Support (ACCESS) program, which is supported by the
NSF grants \#2138259, \#2138286, \#2138307, \#2137603, and \#2138296.
Any opinions, findings, and conclusions or recommendations expressed
in this work are
those of the author(s) and do not necessarily reflect the views of
the NSF.

\bibliography{report} 

@article{Jack2008TheAD,
  title   = {The Alzheimer's disease neuroimaging initiative (ADNI): MRI methods},
  author  = {Clifford R. Jack and Matt A. Bernstein and Nick C Fox and Paul M. Thompson and Gene E. Alexander and Danielle J. Harvey and Bret J. Borowski and Paula J. Britson and Jennifer L Whitwell and Chadwick P. Ward and Anders M. Dale and Joel P. Felmlee and Jeffrey L. Gunter and Derek L. G. Hill and Ronald J. Killiany and Norbert Schuff and Sabrina Fox‐Bosetti and Chen Lin and Colin Studholme and Charles DeCarli and Gunnar Krueger and Heidi A. Ward and Gregory J. Metzger and Katherine T. Scott and Richard Philip Mallozzi and Daniel J. Blezek and Joshua Levy and Josef P. Debbins and Adam Fleisher and Marilyn S. Albert and Robert C. Green and George Bartzokis and Gary H. Glover and John P. Mugler and Michael Weiner},
  journal = {Journal of Magnetic Resonance Imaging},
  year    = {2008},
  volume  = {27},
  doi     = {https://doi.org/10.1002/jmri.21049}
}

@incollection{GRANZIERA2015917,
  title     = {Brain Inflammation, Degeneration, and Plasticity in Multiple Sclerosis},
  editor    = {Arthur W. Toga},
  booktitle = {Brain Mapping},
  publisher = {Academic Press},
  address   = {Waltham},
  pages     = {917-927},
  year      = {2015},
  isbn      = {978-0-12-397316-0},
  doi       = {https://doi.org/10.1016/B978-0-12-397025-1.00109-3},
  url       = {https://www.sciencedirect.com/science/article/pii/B9780123970251001093},
  author    = {C. Granziera and T. Sprenger}
}

@article{Esteban2018,
  title     = {fMRIPrep: a robust preprocessing pipeline for functional MRI},
  volume    = {16},
  issn      = {1548-7105},
  url       = {http://dx.doi.org/10.1038/s41592-018-0235-4},
  doi       = {10.1038/s41592-018-0235-4},
  number    = {1},
  journal   = {Nature Methods},
  publisher = {Springer Science and Business Media LLC},
  author    = {Esteban,  Oscar and Markiewicz,  Christopher J. and Blair,  Ross W. and Moodie,  Craig A. and Isik,  A. Ilkay and Erramuzpe,  Asier and Kent,  James D. and Goncalves,  Mathias and DuPre,  Elizabeth and Snyder,  Madeleine and Oya,  Hiroyuki and Ghosh,  Satrajit S. and Wright,  Jessey and Durnez,  Joke and Poldrack,  Russell A. and Gorgolewski,  Krzysztof J.},
  year      = {2018},
  month     = dec,
  pages     = {111-116}
}

@article{Gorgolewski2016TheBI,
  title   = {The brain imaging data structure, a format for organizing and describing outputs of neuroimaging experiments},
  author  = {Krzysztof J. Gorgolewski and Tibor Auer and Vince D. Calhoun and Richard Cameron Craddock and Samir Das and Eugene P. Duff and Guillaume Flandin and Satrajit S. Ghosh and Tristan Glatard and Yaroslav O. Halchenko and Daniel A. Handwerker and Michael Hanke and David B. Keator and Xiangrui Li and Zachary Michael and Camille Maumet and B. Nolan Nichols and Thomas E. Nichols and John Pellman and Jean-Baptiste Poline and Ariel S. Rokem and Gunnar Schaefer and Vanessa V. Sochat and William Triplett and Jessica A. Turner and Ga{\"e}l Varoquaux and Russell A. Poldrack},
  journal = {Scientific Data},
  year    = {2016},
  volume  = {3}
}

@article{Mulumba2025,
  title     = {The role of neuroimaging in Alzheimer’s disease: implications for the diagnosis,  monitoring disease progression,  and treatment},
  volume    = {4},
  issn      = {2834-5347},
  url       = {http://dx.doi.org/10.37349/en.2025.100675},
  doi       = {10.37349/en.2025.100675},
  journal   = {Exploration of Neuroscience},
  publisher = {Open Exploration Publishing},
  author    = {Mulumba,  Julius and Duan,  Rui and Luo,  Bo and Wu,  Jiang and Sulaiman,  Muhammad and Wang,  Feng and Yang,  Yong},
  year      = {2025},
  month     = feb
}

@article{Schaefer2017,
  title     = {Local-Global Parcellation of the Human Cerebral Cortex from Intrinsic Functional Connectivity MRI},
  volume    = {28},
  issn      = {1460-2199},
  url       = {http://dx.doi.org/10.1093/cercor/bhx179},
  doi       = {10.1093/cercor/bhx179},
  number    = {9},
  journal   = {Cerebral Cortex},
  publisher = {Oxford University Press (OUP)},
  author    = {Schaefer,  Alexander and Kong,  Ru and Gordon,  Evan M and Laumann,  Timothy O and Zuo,  Xi-Nian and Holmes,  Avram J and Eickhoff,  Simon B and Yeo,  B T Thomas},
  year      = {2017},
  month     = jul,
  pages     = {3095–3114}
}

@inproceedings{moghaddami25a,
  title     = {Transformer Model for Alzheimer's Disease Progression Prediction Using  Longitudinal Visit Sequences},
  author    = {Moghaddami, Mahdi and Schubring, Clayton and Siadat, Mohammad},
  booktitle = {Proceedings of the sixth Conference on Health, Inference, and Learning},
  pages     = {804--816},
  year      = {2025},
  editor    = {Xu, Xuhai Orson and Choi, Edward and Singhal, Pankhuri and Gerych, Walter and Tang, Shengpu and Agrawal, Monica and Subbaswamy, Adarsh and Sizikova, Elena and Dunn, Jessilyn and Daneshjou, Roxana and Sarker, Tasmie and McDermott, Matthew and Chen, Irene},
  volume    = {287},
  series    = {Proceedings of Machine Learning Research},
  month     = {25--27 Jun},
  publisher = {PMLR},
  url       = {https://proceedings.mlr.press/v287/moghaddami25a.html}
}

@article{Li2020BrainGNNIB,
  title   = {BrainGNN: Interpretable Brain Graph Neural Network for fMRI Analysis},
  author  = {Xiaoxiao Li and Yuan Zhou and Siyuan Gao and Nicha C. Dvornek and Muhan Zhang and Juntang Zhuang and Shi Gu and Dustin Scheinost and Lawrence H. Staib and Pamela Ventola and James S. Duncan},
  journal = {bioRxiv},
  year    = {2020}
}

@article{Hamilton2017InductiveRL,
  title   = {Inductive Representation Learning on Large Graphs},
  author  = {William L. Hamilton and Zhitao Ying and Jure Leskovec},
  journal = {ArXiv},
  year    = {2017},
  volume  = {abs/1706.02216},
  url     = {https://api.semanticscholar.org/CorpusID:4755450}
}

@inproceedings{Cai2020GraphNormAP,
  title     = {GraphNorm: A Principled Approach to Accelerating Graph Neural Network Training},
  author    = {Tianle Cai and Shengjie Luo and Keyulu Xu and Di He and Tie-Yan Liu and Liwei Wang},
  booktitle = {International Conference on Machine Learning},
  year      = {2020},
  url       = {https://api.semanticscholar.org/CorpusID:221516279}
}

@article{hochreiter1997long,
  title     = {Long short-term memory},
  author    = {Hochreiter, Sepp and Schmidhuber, J{\"u}rgen},
  journal   = {Neural computation},
  volume    = {9},
  number    = {8},
  pages     = {1735--1780},
  year      = {1997},
  publisher = {MIT Press}
}

@inproceedings{Cho2014LearningPR,
  title     = {Learning Phrase Representations using RNN Encoder–Decoder for Statistical Machine Translation},
  author    = {Kyunghyun Cho and Bart van Merrienboer and Çaglar G{\"u}lçehre and Dzmitry Bahdanau and Fethi Bougares and Holger Schwenk and Yoshua Bengio},
  booktitle = {Conference on Empirical Methods in Natural Language Processing},
  year      = {2014}
}

@article{Lin2017FocalLF,
  title   = {Focal Loss for Dense Object Detection},
  author  = {Tsung-Yi Lin and Priya Goyal and Ross B. Girshick and Kaiming He and Piotr Doll{\'a}r},
  journal = {2017 IEEE International Conference on Computer Vision (ICCV)},
  year    = {2017},
  pages   = {2999-3007},
  url     = {https://api.semanticscholar.org/CorpusID:47252984}
}

@article{Bradley1997TheUO,
  title   = {The use of the area under the ROC curve in the evaluation of machine learning algorithms},
  author  = {Andrew P. Bradley},
  journal = {Pattern Recognit.},
  year    = {1997},
  volume  = {30},
  pages   = {1145-1159}
}

@inproceedings{Kim2021,
  title     = {Interpretable temporal graph neural network for prognostic prediction of Alzheimer’s disease using longitudinal neuroimaging data},
  url       = {http://dx.doi.org/10.1109/bibm52615.2021.9669504},
  doi       = {10.1109/bibm52615.2021.9669504},
  booktitle = {2021 IEEE International Conference on Bioinformatics and Biomedicine (BIBM)},
  publisher = {IEEE},
  author    = {Kim,  Mansu and Kim,  Jaesik and Qu,  Jeffrey and Huang,  Heng and Long,  Qi and Sohn,  Kyung-Ah and Kim,  Dokyoon and Shen,  Li},
  year      = {2021},
  month     = dec,
  pages     = {1381–1384}
}

@misc{https://doi.org/10.48550/arxiv.2412.06212,
  doi       = {10.48550/ARXIV.2412.06212},
  url       = {https://arxiv.org/abs/2412.06212},
  author    = {Wang,  Zhepeng and Bao,  Runxue and Wu,  Yawen and Liu,  Guodong and Yang,  Lei and Zhan,  Liang and Zheng,  Feng and Jiang,  Weiwen and Zhang,  Yanfu},
  title     = {A Self-guided Multimodal Approach to Enhancing Graph Representation Learning for Alzheimer's Diseases},
  publisher = {arXiv},
  year      = {2024},
  copyright = {arXiv.org perpetual,  non-exclusive license}
}

@article{Grammenos2024,
  title     = {Predicting the Conversion from Mild Cognitive Impairment to Alzheimer’s Disease Using an Explainable AI Approach},
  volume    = {15},
  issn      = {2078-2489},
  url       = {http://dx.doi.org/10.3390/info15050249},
  doi       = {10.3390/info15050249},
  number    = {5},
  journal   = {Information},
  publisher = {MDPI AG},
  author    = {Grammenos,  Gerasimos and Vrahatis,  Aristidis G. and Vlamos,  Panagiotis and Palejev,  Dean and Exarchos,  Themis},
  year      = {2024},
  month     = apr,
  pages     = {249}
}

@article{Marinescu2018TADPOLECP,
  title   = {TADPOLE Challenge: Prediction of Longitudinal Evolution in Alzheimer's Disease.},
  author  = {Razvan V. Marinescu and Neil P. Oxtoby and Alexandra L. Young and Esther Bron and Arthur W. Toga and Michael W. Weiner and Frederik Barkhof and Nick C Fox and Stefan Klein and Daniel C. Alexander and the EuroPOND Consortium and For the Alzheimer’s Disease Neuroimaging Initiative},
  journal = {arXiv: Populations and Evolution},
  year    = {2018}
}

@article{SCHELTENS2016505,
  title   = {Alzheimer's disease},
  journal = {The Lancet},
  volume  = {388},
  number  = {10043},
  pages   = {505-517},
  year    = {2016},
  issn    = {0140-6736},
  doi     = {https://doi.org/10.1016/S0140-6736(15)01124-1},
  url     = {https://www.sciencedirect.com/science/article/pii/S0140673615011241},
  author  = {Philip Scheltens and Kaj Blennow and Monique M B Breteler and Bart {de Strooper} and Giovanni B Frisoni and Stephen Salloway and Wiesje Maria {Van der Flier}}
}

@article{lancet,
  author  = {Livingston, Gill and Huntley, Jonathan and Sommerlad, Andrew and Ames, David and Ballard, Clive and Banerjee, Sube and Brayne, Carol and Burns, Alistair and Cohen-Mansfield, Jiska and Cooper, Claudia and Costafreda, Sergi G and Dias, Amit and Fox, Nick and Gitlin, Laura N and Howard, Robert and Kales, Helen C and Kivimäki, Mika and Larson, Eric B and Ogunniyi, Adesola and Orgeta, Vasiliki and Ritchie, Karen and Rockwood, Kenneth and Sampson, Elizabeth L and Samus, Quincy and Schneider, Lon S and Selbæk, Geir and Teri, Linda and Mukadam, Naaheed},
  title   = {Dementia prevention, intervention, and care: 2020 report of
             the Lancet Commission},
  journal = {The Lancet},
  year    = {2020},
  volume  = {396},
  doi     = {10.1016/S0140-6736(20)30367-6}
}

@inproceedings{BayesianOptimization,
  author    = {Snoek, Jasper and Larochelle, Hugo and Adams, Ryan P.},
  title     = {Practical Bayesian optimization of machine learning algorithms},
  year      = {2012},
  publisher = {Curran Associates Inc.},
  address   = {Red Hook, NY, USA},
  booktitle = {Proceedings of the 26th International Conference on Neural Information Processing Systems - Volume 2},
  pages     = {2951–2959},
  numpages  = {9},
  location  = {Lake Tahoe, Nevada},
  series    = {NIPS'12}
}
\bibliographystyle{spiebib} 

\end{document}